\documentclass[conference]{IEEEtran}
\IEEEoverridecommandlockouts

\usepackage{cite}
\usepackage{amsmath,amssymb,amsfonts}
\usepackage{booktabs}
\usepackage{graphicx}
\usepackage{textcomp}
\usepackage{xcolor}
\usepackage{url}
\usepackage{multirow}
\usepackage{balance}

\newcommand{\dataset}{WCID}

\begin{document}

\title{Benchmarking Deep Learning Models for Wheelchair Configuration Classification in Accessible Automated Vehicle Applications}

\author{%
\IEEEauthorblockN{%
Samir Ahmed,
Shaaz Saleem,
Devin Dhami,
Wesley Maia,
Angel Martinez-Sanchez,\\
Parthib Roy,
Jannatul Ferdous Srabonee,
Ahmed Sabbir Arif, and
Ross Greer}
\IEEEauthorblockA{%
University of California Merced\\
}
}

\maketitle

\begin{abstract}
Making automated vehicles accessible requires consideration of vulnerable road users, including wheelchair users. To offer a more tailored experience that supports independence for this group, vehicle design needs to accommodate the wide range of wheelchair configurations. However, wheelchairs are individualized, varying in frame geometry, propulsion, and visual appearance. Moreover, occupants or accessories may obscure visual cues. Existing mobility-aid datasets only label the presence of a wheelchair, rather than describing its specific configuration. To fill this gap, this paper introduces the Wheelchair Configuration Image Dataset (\dataset{}), comprising 467 face-blurred images across 192 capture groups solely for evaluation. Each image is annotated with (i) a two-class label of manual or powered, (ii) a chair subtype, (iii) occupancy, viewpoint, setting, lighting, occlusion, propulsion, and control, with a 273-image validation split and 194-image test split. Three convolutional neural network backbones and a frozen vision-language probe are trained on an external image dataset and evaluated on \dataset{} to measure cross-dataset generalization, while also testing robustness using Gaussian blur. Convolutional models do well on clean images, but performance drops under strong blur, leaving a gap before deployment. Code, split assignments, labels, trained weights, and the face-blurred images is available at: \url{https://github.com/Mi3-Lab/MAID}
\end{abstract}

\section{Introduction}

Automated vehicles have the potential to improve mobility independence for wheelchair users, provided that accessibility is addressed throughout vehicle design rather than treated as a post-production modification~\cite{rojas2022shuttle}. A wheelchair-accessible automated vehicle must support a complete trip including ride hailing, boarding, interior maneuvering, securement, occupant restraint, and egress~\cite{golbabaei2024accessibility,klinich2022safety}. Wheelchair accessibility in automated vehicles is continuing to get attention from industry and research, with initiatives such as the SAE--General Motors CivicProgress Challenge~\cite{sae2026civicprogress} asking university teams to retrofit a vehicle for wheelchair access, including entrance, securement, and adaptive seating systems. Within that context, this study addresses a foundational perception question: \emph{How accurately can visual classifiers trained on external images recover wheelchair configurations from photographs, and what limits their reliability as preliminary inputs for accessible automated vehicles?}

\begin{figure}[!t]
    \centering
    \includegraphics[width=0.92\linewidth]{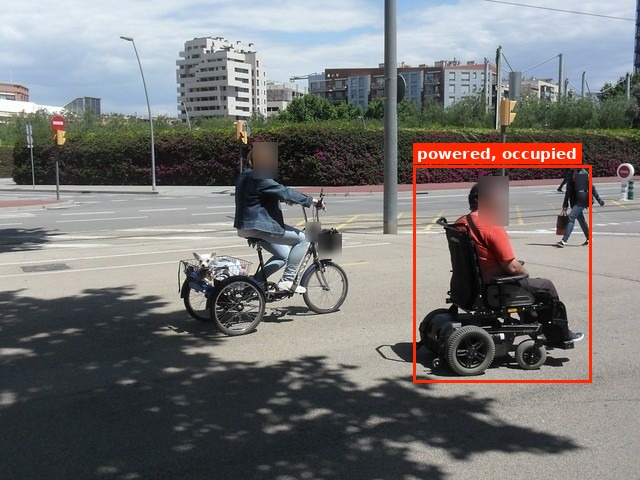}
    \caption{Example sample from \dataset{} (powered, occupied wheelchair in an outdoor street scene with automated face anonymization). Distinguishing powered configurations (via battery housings, drive wheels, and joystick controls) from manual chairs provides early visual context to initialize vehicle boarding and securement checklists. Photo: Eric Fischer, CC BY 2.0.}
    \label{fig:example}
\end{figure}

Wheelchairs are not a visually uniform object class. Manual and powered configurations differ in wheel geometry, frames, seating systems, controls, and batteries, while occupants, bags, and accessories frequently obscure configuration cues (Fig.~\ref{fig:example}). Recovering configuration early in the perception pipeline can help vehicle subsystems anticipate weight and footprint constraints~\cite{klinich2022safety}, and could also initialize an appropriate boarding checklist or prompt human confirmation before compatibility is verified. However, since classifiers trained on web imagery risk memorizing device-specific textures or collection backgrounds, a rigorous evaluation must assess selective deferral and corruption robustness alongside aggregate accuracy on images collected separately from the training data.

This problem is investigated through \dataset{}, an annotated evaluation-only dataset of 467 wheelchair photographs across 192 capture groups. Four established model families are evaluated and trained solely on external images, are evaluated on \dataset{} for classification performance, selective deferral, and blur robustness. This paper contributes three things: (1) a publicly documented, configuration-oriented wheelchair dataset with group and capture-condition metadata; (2) a cross-dataset evaluation of accuracy and selective deferral; and (3) robustness analysis across Gaussian blur. Code, split assignments, labels, trained weights, and face-blurred images will be released (where licenses permit) to support reproducible research.

\section{Related Work}

\subsection{Accessible Automated Vehicles}

Golbabaei \emph{et al.} reviewed 57 studies to create design
requirements for accessible AVs across the whole trip, and identify
acceptance, accessibility, and trust as the central
concerns~\cite{golbabaei2024accessibility}. For independent boarding, they
highlighted ramps no steeper than 1:8, clear floor space sized for the
largest chairs (powered chairs are the limiting case, at 870~mm by
1508~mm), and securement that riders can engage without help, which no
standard yet provides for all wheelchairs. For trust, they recommend trial
periods before large-scale deployment, along with education and inclusive
pilots~\cite{golbabaei2024accessibility}. They do not address how a vehicle
would tell which device is approaching.

Klinich \emph{et al.} conclude that a wheelchair tiedown and occupant
restraint system (WTORS) for AVs must be crashworthy, usable independently,
and adaptable to a wide range of wheelchair types~\cite{klinich2022safety}.
Existing systems do not meet all three, and the proposed Universal Docking
Interface Geometry (UDIG) depends on voluntary standards and compatible
hardware on both the vehicle and the chair~\cite{klinich2022safety}. Until
that hardware is universal, AVs will meet heterogeneous chairs.

Rojas \emph{et al.} simulate shuttles and show that a retrofit for
wheelchair access cut driving range by 38\% and worsened turning radius and
cost per passenger, while a purpose-designed variant did not, so they argue
accessibility belongs in the design phase~\cite{rojas2022shuttle}.

These studies show that boarding space and securement depend on the rider's
chair, but not how the vehicle would learn which configuration it is about
to serve. \dataset{} addresses this perception side by labeling manual
versus powered configuration, subtype, and capture conditions on
photographs held out from training, so a camera-based estimate can be
tested before it is used to prepare boarding or securement. Mass,
dimensions, tiedown integrity, crashworthiness, docking compatibility, and
vehicle dynamics are out of scope. Because an image label cannot confirm
them~\cite{klinich2022safety}, we treat classification as an
uncertainty-aware upstream filter that defers low-confidence predictions,
not as an actuating controller.
\subsection{Visual Recognition of Mobility Aids}

There are few existing datasets, with their primary focus being detection-based~\cite{mohr2023crossroad,liu2026pmma} and PMMA describing its wheelchair class as self-propelled~\cite{liu2026pmma}. Detection labels do not resolve within-wheelchair configuration (e.g., manual vs.\ powered), which governs weight, footprint, and securement preparation~\cite{klinich2022safety}. Even within a propulsion class, subtypes impose different physical constraints: negative-camber manual wheelchairs for sporting are easier to push and turn than standard manual chairs, but their angled wheel geometry and active balance make them easier to tip during sharp rotations or ramp ascent, whereas standard manual frames are harder to push but have more stability against tipping. Similarly, powered tilt-recline wheelchairs introduce articulating seat components, shifting center of gravity, and an expanded set of clearances to consider compared to standard powered chairs. \dataset{} targets this gap by providing configuration-level and capture-condition annotations so models can be evaluated on these visual distinctions.

\begin{table}[!t]
\centering
\caption{WCID Configuration Taxonomy. Two classes separate propulsion mechanisms (manual vs.\ powered) with data to additionally differentiate within the classes.}
\label{tab:taxonomy}
\footnotesize
\begin{tabular}{@{}p{0.27\columnwidth}p{0.68\columnwidth}@{}}
\toprule
\multicolumn{2}{@{}l}{\textbf{Class (prediction target)}} \\
\midrule
Manual  & Wheelchairs with no integrated electric propulsion; self-propelled via hand rims on large rear wheels, or attendant-propelled. \\
\addlinespace[2pt]
Powered & Wheelchairs with electric propulsion; visible drive wheels, battery housing, joystick or similar control. \\
\midrule
\multicolumn{2}{@{}l}{\textbf{Subtype (descriptive)}} \\
\midrule
\multicolumn{2}{@{}l}{\emph{Manual}} \\
\quad Standard            & Conventional frame. \\
\quad Negative camber     & Rear wheels angled inward at the top. \\
\quad Transport/attendant & Small rear wheels, no hand rims, push handles. \\
\quad Tilt-recline        & Seat or backrest tilts or reclines. \\
\quad Bariatric           & Visibly wider, heavier-duty frame. \\
\multicolumn{2}{@{}l}{\emph{Powered}} \\
\quad Standard            & Conventional powered chair. \\
\quad Tilt-recline        & Powered seat tilt or recline. \\
\multicolumn{2}{@{}l}{\emph{Either class}} \\
\quad Unknown             & Subtype not determinable from the image. \\
\midrule
\multicolumn{2}{@{}l}{\textbf{Per-image attributes (descriptive)}} \\
\midrule
Occupancy  & occupied, unoccupied \\
Viewpoint  & front, rear, side, three-quarter \\
Setting    & curb, vehicle, sidewalk, indoor, other \\
Lighting   & daylight, shade, indoor, low light \\
Occlusion  & none, partial, heavy \\
Propulsion & self, attendant, not applicable, unclear \\
Control    & joystick, attendant control, push rims, tiller, none visible, unknown \\
\bottomrule
\end{tabular}
\end{table}

\subsection{Selective Prediction, Robustness, and Dataset Shift}

Because the classifier serves as a deferrable perception input, it does not need to force a prediction on every frame: misclassifying a wheelchair's configuration risks initiating an unsupported entrance or securement process, whereas abstaining safely defers to an alternative camera view or human validation. Selective classification allows a classifier to abstain when presented with low-confidence inputs, trading coverage for smaller error on accepted inputs. Ranking inputs by maximum softmax probability (referred to as softmax response) is a strong confidence measure for deep networks, and the trade-off is summarized by the risk--coverage curve~\cite{geifman2017selective}. Images can be degraded under unfavorable conditions, requiring testing of convolutional neural networks under Gaussian blurring~\cite{dodge2016quality,hendrycks2019imagenetc}. Datasets carry their own biases, causing models to lose accuracy when tested on separate datasets from their training set~\cite{recht2019imagenet,torralba2011bias}. \dataset{} addresses these gaps by providing a held-out, group-exclusive evaluation benchmark which jointly measures cross-dataset generalization, selective deferral, and blur robustness.

\subsection{Candidate Classifier Architectures}
Evaluating wheelchair configuration recognition for automated vehicles requires comparing architectures across spatial discrimination and out-of-distribution resilience. To span the convolutional model space, three ImageNet-pretrained backbones are benchmarked: ResNet-50~\cite{he2016resnet} serves as a standard residual baseline for visual transfer learning; EfficientNet-B0~\cite{tan2019efficientnet} represents scaled mobile architectures for edge deployment; and ConvNeXt-Tiny~\cite{liu2022convnext} modernizes standard ConvNets with $7\times7$ depthwise convolutions which capture spatial context across fine-grained categories and domain shifts. Because end-to-end fine-tuning on external images can distort pretrained features on out-of-distribution targets~\cite{kumar2022finetuningdistortpretrainedfeatures}, a frozen SigLIP~2~\cite{tschannen2025siglip2} vision-language encoder is also evaluated with a linear probe. Comparing fine-tuned CNNs against a frozen vision-language pretrained representation clarifies how architecture and adaptation strategy affect accuracy, selective deferral, and blur sensitivity on \dataset{}.
\section{Dataset and Benchmark Methodology}

\subsection{Dataset Scope, Taxonomy, and Quality Control}

\begin{figure}[t]
\centering
{\footnotesize Manual}\\[2pt]
\begin{minipage}[c]{0.48\columnwidth}\centering
\includegraphics[width=\linewidth,height=4.5cm,keepaspectratio]{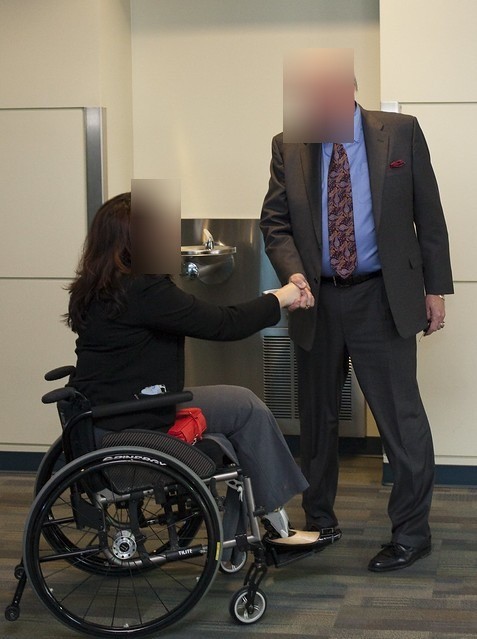}\\[1pt]
{\scriptsize (a) (manual, test): occupied, self-propelled, side view, indoor, partial occlusion, standard}
\end{minipage}\hfill
\begin{minipage}[c]{0.48\columnwidth}\centering
\includegraphics[width=\linewidth,height=4.5cm,keepaspectratio]{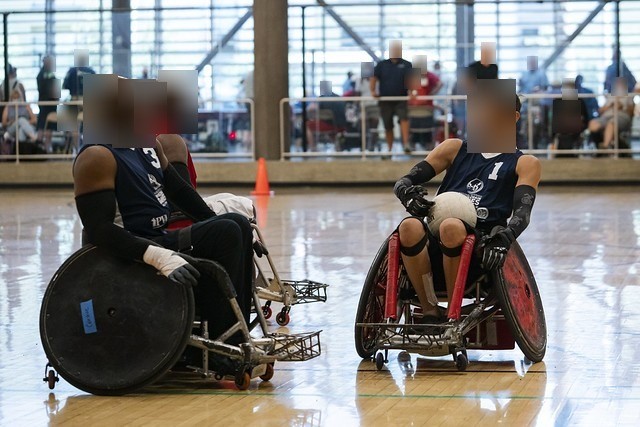}\\[1pt]
{\scriptsize (b) (manual, val): occupied, self-propelled, side view, indoor, no occlusion, negative-camber}
\end{minipage}
\\[6pt]
{\footnotesize Powered}\\[2pt]
\begin{minipage}[c]{0.48\columnwidth}\centering
\includegraphics[width=\linewidth,height=4.5cm,keepaspectratio]{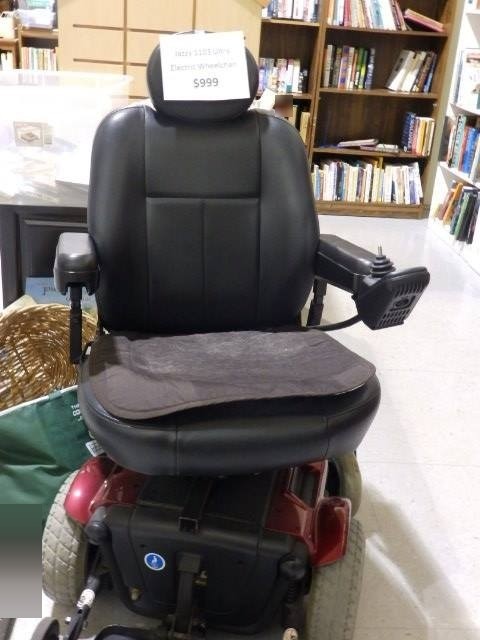}\\[1pt]
{\scriptsize (c) (powered, test): unoccupied, front view, indoor, joystick visible, partial occlusion, power standard}
\end{minipage}\hfill
\begin{minipage}[c]{0.48\columnwidth}\centering
\includegraphics[width=\linewidth,height=4.5cm,keepaspectratio]{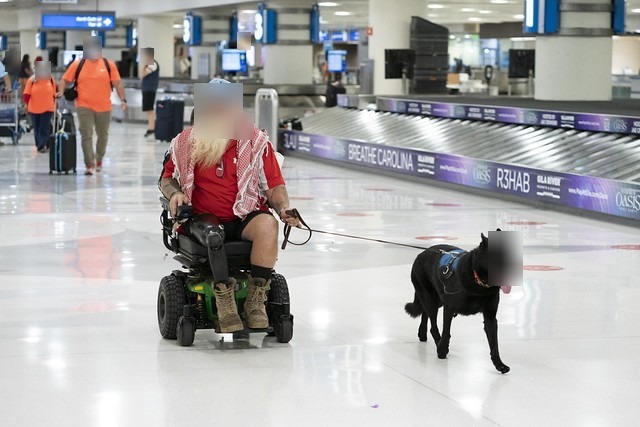}\\[1pt]
{\scriptsize (d) (powered, val): occupied, front view, indoors, joystick visible, no occlusion, power standard}
\end{minipage}
\vspace{6pt}
\caption{Example WCID images; top row manual, bottom row powered. Faces are blurred where present. Image credits (left to right, top to bottom): COD Newsroom (CC~BY~2.0); U.S.\ Dept.\ of Veterans Affairs (public domain); Rikki's Refuge (CC~BY~2.0); U.S.\ Dept.\ of Veterans Affairs (public domain).}
\label{fig:wcid_examples}
\end{figure}

\dataset{} is curated to study wheelchair configuration recognition across variations in device geometry, rider occupancy, and capture environments. Table~\ref{tab:taxonomy} defines the two-class taxonomy separating manual and powered wheelchairs and the finer data that differentiates certain subtypes. Each sample also records occupancy (occupied or unoccupied) and scene metadata (viewpoint, setting, lighting, occlusion, propulsion, and control)~(Fig.~\ref{fig:wcid_examples}). To protect privacy in third-party photographs, all detected faces are anonymized via pixelation and Gaussian blur prior to release.

Annotations were restricted to directly observable visual features. Table~\ref{tab:kappa} reports inter-annotator agreement (Cohen's $\kappa$) across doubly annotated samples. High agreement on occupancy ($\kappa=1.00$), device type ($\kappa=0.86$), propulsion ($\kappa=0.76$), and lighting ($\kappa=0.73$) suggests that these attributes support reliable labels, whereas occlusion ($\kappa=0.05$) displayed near-chance agreement. During quality control, 274 candidate web images were excluded due to non-photographic content (146), child subjects (116), unblurred identifiable faces (7), licensing restrictions (3), or clinical backgrounds (2). Seventy-five test images from PMMA~\cite{liu2026pmma} were excluded after auditing revealed systematic taxonomy mismatches in which powered consensus labels clashed with PMMA's self-propelled description.

\begin{table}[!t]
\centering
\caption{Inter-annotator agreement (Cohen's $\kappa$) over doubly annotated images ($n$). High agreement on occupancy, device type, propulsion, and lighting supports their use as evaluation labels, whereas occlusion is around $\kappa = 0$.}
\label{tab:kappa}
\footnotesize
\begin{tabular}{@{}lcr@{\hspace{18pt}}lcr@{}}
\toprule
Field & Cohen's $\kappa$ & $n$ & Field & Cohen's $\kappa$ & $n$ \\
\midrule
Occupancy   & 1.00 & 102 & Setting   & 0.59 & 102 \\
Device type & 0.86 & 102 & Viewpoint & 0.45 & 102 \\
Propulsion  & 0.76 & 102 & Occlusion & 0.05 & 102 \\
Lighting    & 0.73 & 102 &           &      &     \\
\bottomrule
\end{tabular}
\end{table}

\subsection{Group-Exclusive Partitioning}
\label{sec:partitioning}

Random image-level splitting inflates generalization metrics when multiple photographs capture the same physical wheelchair or originate from a single web listing. Group-exclusive partitioning is therefore enforced: every image is assigned a composite group key combining its creator/uploader, source page, Flickr photo ID, capture burst, title series, and near-duplicate cluster (perceptual hash and SigLIP 2 embedding similarity). All images sharing a group key are restricted to a single split (Table~\ref{tab:splits}). The two-class benchmark (467 images, 192 groups from OpenImages and Openverse) is partitioned into 273 validation and 194 test images (83 and 109 groups, respectively), with training run on 4,362 external images that never enter validation or test splits.

\subsection{Evaluation, Robustness, and Model Selection Methodology}

For $C$ classes, classification performance is quantified
using overall accuracy
($\mathrm{Accuracy}=\frac{1}{N}\sum_{c=1}^{C}TP_c$),
balanced accuracy, per-class recall, and macro-averaged
F1, defined as
\begin{equation}
\mathrm{MacroF1} =
\frac{1}{C}\sum_{c=1}^{C}
\frac{2P_cR_c}{P_c+R_c}.
\label{eq:macro_f1}
\end{equation}
Here, $N$ is the total number of samples,
$TP_c$ denotes true positives for class $c$,
and $P_c$ and $R_c$ denote its precision and recall,
respectively. Macro-F1 is the primary discrimination
metric under class imbalance.

Selective prediction is evaluated via the area under the
risk--coverage curve (AURC) and retained accuracy at
90\%, 80\%, and 70\% coverage, ranking predictions by
maximum softmax probability~\cite{geifman2017selective}.

To probe sensitivity to high-frequency spatial cues (fine structures such as wheel spokes, hand rims, and joystick boxes), trained models are evaluated on clean test images ($\sigma=0$) versus images degraded by isotropic Gaussian blur $I_{\sigma}=G_{\sigma}*I$ across $\sigma\in\{0.5,1,2,4\}$ pixels under two regimes: (1) \emph{Blur before resize}, applied after capping the longer side at 640 pixels prior to downsampling ($\sigma$ in pixels of the 640-capped image), and (2) \emph{Blur after resize}, applied directly at native model-input resolution ($\sigma$ in model-input pixels, 224 for the CNNs and 384 for SigLIP 2).

The validation split is used only for checkpoint selection (by validation macro-F1), and the test split is evaluated post-selection for each model and seed on clean images.

\begin{table}[!t]
\caption{WCID splits (seed 20261111). Models are trained on 4,362 external images only. OI: OpenImages; OV: Openverse.}
\label{tab:splits}
\centering
\footnotesize
\begin{tabular}{@{}lrrll@{}}
\toprule
Split & Img. & Groups & Class counts & Image source \\
\midrule
Val  & 273 & 83  & manual 207, powered 66  & OI 86, OV 187 \\
Test & 194 & 109 & manual 147, powered 47  & OI 176, OV 18 \\
All  & 467 & 192 & manual 354, powered 113 & OI 262, OV 205 \\
\bottomrule
\end{tabular}
\end{table}

\subsection{Implementation Details}

Four pretrained architectures are benchmarked: \textbf{ResNet-50}~\cite{he2016resnet}, \textbf{EfficientNet-B0}~\cite{tan2019efficientnet}, \textbf{ConvNeXt-Tiny}~\cite{liu2022convnext}, and a frozen \textbf{SigLIP~2} (\texttt{so400m-patch14-384}) vision--language encoder paired with a linear probe~\cite{tschannen2025siglip2}. Images are decoded with a 640-pixel longest-side cap and resized to $224\times224$ for the three CNNs and $384\times384$ for SigLIP~2. The CNNs are initialized from ImageNet weights and fine-tuned end-to-end, whereas SigLIP~2 serves as a frozen-feature baseline, with the head trained on features that were cached on unaugmented images. CNN training applies geometry-preserving augmentations: random resized cropping (scale $0.7\text{--}1.0$), horizontal flipping ($p=0.5$), color jitter, and minor affine transforms ($\pm 8^\circ$ rotation, $5\%$ translation, $0.95\text{--}1.05$ scale). Each model also has an auxiliary occupied/unoccupied head, with its cross-entropy added with weight $1.0$ on training images with an occupancy label, but only the manual vs. powered output is used for selection and evaluation. All models are optimized in mixed precision using AdamW (learning rate $10^{-4}$, weight decay $10^{-4}$, batch size 64), cosine annealing, and class-weighted cross-entropy for up to 50 epochs with 8-epoch early stopping on validation macro-F1.

\begin{table*}[!t]
\centering
\caption{Two-class test results for the four models. Run on the 194-image test split, 109 groups; mean $\pm$ s.d.\ over seeds 0--2; CI: 95\% group-bootstrap interval (1,000 resamples of test groups, mean of per-seed endpoints); best results bolded; SigLIP~2 is listed separately because it is a frozen backbone with a linear head.}
\label{tab:main-results}
\footnotesize
\setlength{\tabcolsep}{4pt}
\begin{tabular}{@{}lccccccc@{}}
\toprule
Model & Acc. & Acc.\ 95\% CI & Bal.\ acc. & Macro-F1 & F1 95\% CI & Manual Recall & Powered Recall \\
\midrule
ResNet-50       & $0.878{\pm}0.044$ & [0.818, 0.929] & $0.845{\pm}0.038$ & $0.838{\pm}0.046$ & [0.761, 0.901] & $0.909{\pm}0.081$ & $0.780{\pm}0.117$ \\
EfficientNet-B0 & $0.835{\pm}0.034$ & [0.774, 0.892] & $0.787{\pm}0.069$ & $0.779{\pm}0.056$ & [0.697, 0.846] & $0.880{\pm}0.021$ & $0.695{\pm}0.142$ \\
ConvNeXt-Tiny   & $\mathbf{0.921{\pm}0.006}$ & [0.883, 0.956] & $\mathbf{0.892{\pm}0.004}$ & $\mathbf{0.892{\pm}0.007}$ & [0.833, 0.937] & $\mathbf{0.948{\pm}0.010}$ & $\mathbf{0.837{\pm}0.012}$ \\
\midrule
SigLIP 2 (frozen) & $0.756{\pm}0.017$ & [0.681, 0.823] & $0.706{\pm}0.017$ & $0.690{\pm}0.018$ & [0.605, 0.770] & $0.803{\pm}0.018$ & $0.610{\pm}0.025$ \\
\bottomrule
\end{tabular}
\end{table*}

\begin{table*}[!t]
\centering
\caption{Post-hoc two-class macro-F1 before and after resizing on the 194-image test split under Gaussian blur; mean $\pm$ s.d.\ over seeds 0--2; best per regime and $\sigma$ bolded.}
\label{tab:blur-before-after}
\footnotesize
\begin{tabular}{@{}llccccc@{}}
\toprule
Model & Blur & $\sigma{=}0$ & $\sigma{=}0.5$ & $\sigma{=}1$ & $\sigma{=}2$ & $\sigma{=}4$ \\
\midrule
ResNet-50         & Before & $0.838{\pm}0.046$ & $0.831{\pm}0.023$ & $0.821{\pm}0.026$ & $0.805{\pm}0.028$ & $0.715{\pm}0.017$ \\
ResNet-50         & After  & $0.838{\pm}0.046$ & $0.822{\pm}0.023$ & $0.808{\pm}0.027$ & $0.712{\pm}0.020$ & $0.522{\pm}0.099$ \\
EfficientNet-B0   & Before & $0.779{\pm}0.056$ & $0.780{\pm}0.061$ & $0.777{\pm}0.056$ & $0.740{\pm}0.032$ & $0.636{\pm}0.083$ \\
EfficientNet-B0   & After  & $0.779{\pm}0.056$ & $0.785{\pm}0.061$ & $0.748{\pm}0.025$ & $0.618{\pm}0.075$ & $0.476{\pm}0.018$ \\
ConvNeXt-Tiny     & Before & $\mathbf{0.892{\pm}0.007}$ & $\mathbf{0.889{\pm}0.001}$ & $\mathbf{0.857{\pm}0.013}$ & $\mathbf{0.814{\pm}0.015}$ & $\mathbf{0.756{\pm}0.032}$ \\
ConvNeXt-Tiny     & After  & $\mathbf{0.892{\pm}0.007}$ & $\mathbf{0.866{\pm}0.010}$ & $\mathbf{0.820{\pm}0.028}$ & $\mathbf{0.739{\pm}0.032}$ & $\mathbf{0.598{\pm}0.055}$ \\
\midrule
SigLIP~2 (frozen) & Before & $0.690{\pm}0.018$ & $0.680{\pm}0.017$ & $0.661{\pm}0.014$ & $0.606{\pm}0.024$ & $0.567{\pm}0.016$ \\
SigLIP~2 (frozen) & After  & $0.690{\pm}0.018$ & $0.688{\pm}0.019$ & $0.638{\pm}0.014$ & $0.603{\pm}0.013$ & $0.529{\pm}0.020$ \\
\bottomrule
\end{tabular}
\end{table*}

\section{Results}

\subsection{Classification}

Table~\ref{tab:main-results} presents locked two-class test performance (194 images, 109 groups). The three fine-tuned CNNs achieve strong discrimination: mean accuracy spans $0.835\text{--}0.921$, balanced accuracy $0.787\text{--}0.892$, and macro-F1 $0.779\text{--}0.892$. ConvNeXt-Tiny attains the highest mean accuracy ($0.921\pm0.006$), macro-F1 ($0.892\pm0.007$), manual recall ($0.948\pm0.010$), and powered recall ($0.837\pm0.012$). All three CNNs' bootstrap 95\% confidence intervals overlap: ConvNeXt's F1 CI is [0.833, 0.937], EfficientNet's is [0.697, 0.846], and ResNet's is [0.761, 0.901]. Every model achieved lower powered than manual recall.

\begin{table}[!t]
\centering
\caption{Results for selective deferral on the 194-image test split, 109 groups; confidence = maximum softmax probability; mean $\pm$ s.d.\ over seeds 0--2.}
\label{tab:deferral}
\resizebox{\columnwidth}{!}{%
\begin{tabular}{@{}lcccc@{}}
\toprule
Model & AURC $\downarrow$ & Acc.@90\% $\uparrow$ & Acc.@80\% $\uparrow$ & Acc.@70\% $\uparrow$ \\
\midrule
ResNet-50       & $0.026{\pm}0.010$ & $0.909{\pm}0.030$ & $0.940{\pm}0.016$ & $0.968{\pm}0.024$ \\
EfficientNet-B0 & $0.043{\pm}0.014$ & $0.874{\pm}0.032$ & $0.912{\pm}0.032$ & $0.946{\pm}0.021$ \\
ConvNeXt-Tiny   & $\mathbf{0.015{\pm}0.004}$ & $\mathbf{0.937{\pm}0.010}$ & $\mathbf{0.966{\pm}0.004}$ & $\mathbf{0.980{\pm}0.008}$ \\
\midrule
SigLIP~2 (frozen) & $0.088{\pm}0.001$ & $0.783{\pm}0.010$ & $0.806{\pm}0.006$ & $0.850{\pm}0.004$ \\
\bottomrule
\end{tabular}%
}
\end{table}

\subsection{Selective Deferral}

Maximum softmax probability is used to defer uncertain inputs (Table~\ref{tab:deferral}); ConvNeXt-Tiny achieves the lowest risk--coverage AURC ($0.015\pm0.004$). Retaining the most confident 90\% ($k=175$), 80\% ($k=155$), and 70\% ($k=136$) of the 194 test images increases ConvNeXt-Tiny accuracy from $0.921\pm0.006$ to $0.937\pm0.010$, $0.966\pm0.004$, and $0.980\pm0.008$, respectively.

\subsection{Gaussian Blur Robustness}

\begin{figure}[!t]
    \centering
    \includegraphics[width=\columnwidth]{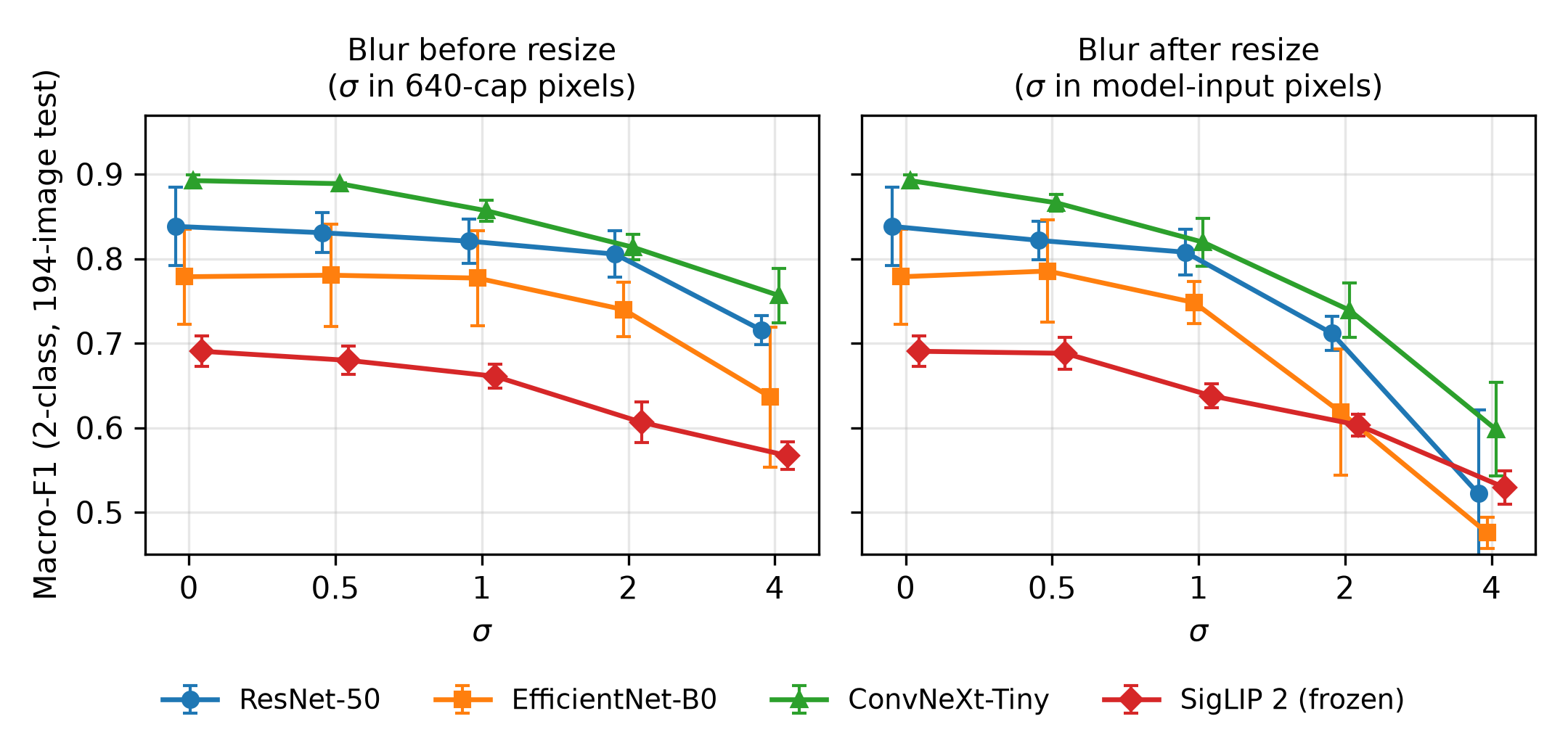}
    \caption{Post-hoc two-class macro-F1 on the 194-image test split across Gaussian blur severity $\sigma$ (mean $\pm$ s.d.\ over seeds 0--2). For the CNNs, blur applied at model-input resolution (right) degrades recognition much more sharply than blur applied before downsampling (left).}
    \label{fig:blur}
\end{figure}

Fig.~\ref{fig:blur} plots macro-F1 on the 194-image test split under a post-hoc Gaussian blur sweep ($\sigma \in \{0.5, 1, 2, 4\}$), with blur applied before resizing (left) and after resizing (right), presented in Table~\ref{tab:blur-before-after}. In both regimes, macro-F1 decreases as $\sigma$ increases, for all models at $\sigma \geq 1$. With blur before resizing, ConvNeXt-Tiny declines from $0.892$ to $0.889$, $0.857$, $0.814$, and $0.756$ at $\sigma = 0.5$, $1$, $2$, and $4$, while ResNet-50 and EfficientNet-B0 reach $0.715$ and $0.636$ at $\sigma = 4$. With blur after resizing, the CNNs decline more sharply: at $\sigma = 4$, ConvNeXt-Tiny falls to $0.598\pm0.055$, ResNet-50 $0.522\pm0.099$, and EfficientNet-B0 $0.476\pm0.018$. SigLIP~2 shows the greatest stability after resizing, dropping from $0.690$ to $0.567\pm0.016$ (before resizing) and $0.529\pm0.020$ (after resizing) at $\sigma = 4$.

\section{Discussion}

\subsection{Toward a Vehicle Interface}

In a conceptual vehicle architecture, an exterior or cabin camera could capture an image of an approaching passenger's and process it through a classification model, which outputs a device-class prediction and maximum softmax probability to an accessibility controller. Rather than directly actuating mechanical systems, high confidence predictions could support adaptive human-machine interfaces (HMI) and proactive cabin preparation to improve the passenger's accessibility and independence. For instance, distinguishing powered and specialized devices from standard manual chairs can identify accommodation needs and prompt the system to offer a compatible in-cabin interaction mode, such as voice-based assistance, hands-free cabin controls, or accessible interface configurations. Classification outputs could also inform preliminary cabin layout recommendations, guide visual instructions for wheelchair securement or docking, and identify where additional geometric measurements may be required. When prediction confidence is low, the system could instead request an alternative camera viewpoint or human confirmation before proceeding. ConvNeXt-Tiny has the highest test macro-F1 and lowest AURC ($0.015\pm0.004$) on the 194-image test split, and its seed spread is the tightest, though its confidence intervals still overlap the nearest model on macro-F1 (ResNet-50). Deferring the least confident 30\% of ConvNeXt-Tiny predictions raises accuracy on the retained 70\% from $0.921\pm0.006$ to $0.980\pm0.008$ (Table~\ref{tab:deferral}), which is consistent with selective deferral as an upstream filter. Only after independent geometric, occupancy, and safety verification could a certified subsystem execute physical boarding or securement actions.

\subsection{Limitations and Future Work}

This benchmark is bounded by dataset scale, web-source bias, and the absence of physical device serial numbers. Groups come from source metadata and image similarity, so the same wheelchair from different uploaders can land in both the test and validation splits. The test set consists of 194 images in 109 groups (147 manual, 47 powered), and macro-F1 confidence intervals span 0.10--0.17, so differences between models are uncertain. Seventy-five PMMA test images were removed after the fact because their powered labels conflicted with PMMA's own self-propelled description. Label agreement of device type was measured on 102 images by two annotators, with Cohen's $\kappa = 0.86$ on this single annotator pair. Powered recall sits below manual recall for every model. Faces are blurred in the 467 \dataset{} images; however, face-blur of external training images is not verified, and plates and signs are not blurred. Closed-set softmax classifiers can assign high confidence to out-of-distribution mobility aids. Future work will expand multi-site physical device capture, co-design taxonomies with wheelchair users, and integrate open-set out-of-distribution detection.

\section{Conclusion}
We introduced \dataset{} and a benchmark for wheelchair device-class classification in accessible automated vehicle applications, evaluating two-class classification on group-exclusive splits, selective deferral, and blur sensitivity across four architectures. On the 194-image test split, ConvNeXt-Tiny achieved the highest mean accuracy ($0.921\pm0.006$) and the lowest selective risk (AURC $0.015\pm0.004$), reaching $0.980\pm0.008$ accuracy when deferring the 30\% least confident predictions. These findings suggest that uncertainty-aware device-class recognition can serve as a contextual input for accessible automated vehicles.


\balance
\bibliographystyle{IEEEtran}
\bibliography{refs}

\end{document}